\documentclass{article} 
\usepackage[table,xcdraw]{xcolor} 
\usepackage{iclr2025_conference,times}

\usepackage{amsmath,amsfonts,bm}

\def\eqref#1{equation~\ref{#1}}

\def\1{\bm{1}}

\DeclareMathAlphabet{\mathsfit}{\encodingdefault}{\sfdefault}{m}{sl}
\SetMathAlphabet{\mathsfit}{bold}{\encodingdefault}{\sfdefault}{bx}{n}

\usepackage{url}
\usepackage{hyperref}
\definecolor{myGreen}{rgb}{0, .6, .0}
\hypersetup{colorlinks,breaklinks,anchorcolor=darkblue,citecolor=myGreen}
\usepackage{multirow}
\usepackage{booktabs}
\usepackage{graphicx}
\usepackage{subfigure}
\usepackage{caption}
\usepackage{subcaption}
\usepackage{makecell}
\usepackage[capitalize]{cleveref}
\usepackage{multicol}
\crefname{section}{Sec.}{Secs.}
\Crefname{section}{Section}{Sections}
\Crefname{table}{Table}{Tables}
\crefname{table}{Tab.}{Tabs.}
\crefname{figure}{Fig.}{Figs.}

\definecolor{deepblue}{RGB}{0,0,180}

\title{
  PruneVid: Visual Token Pruning for Efficient Video Large Language Models
}

\author{{Xiaohu Huang}\textsuperscript{1}\qquad\qquad
{Hao Zhou}\textsuperscript{2} \qquad\qquad {Kai Han}\textsuperscript{1}\thanks{Corresponding author.}\\
\textsuperscript{1} Visual AI Lab, The University of Hong Kong \\
\textsuperscript{2} Department of Computer Vision Technology (VIS), Baidu Inc. \\
{\tt\small huangxiaohu@connect.hku.hk \ \  zhouh156@mail.ustc.edu.cn \ \  kaihanx@hku.hk}
}

\iclrfinalcopy 
\begin{document}

\maketitle

\begin{abstract}

In this paper, we introduce PruneVid, a visual token pruning method designed to enhance the efficiency of multi-modal video understanding. Large Language Models (LLMs) have shown promising performance in video tasks due to their extended capabilities in comprehending visual modalities. However, the substantial redundancy in video data presents significant computational challenges for LLMs. To address this issue, we introduce a training-free method that 1) minimizes video redundancy by merging spatial-temporal tokens, and 2) leverages LLMs’ reasoning capabilities to selectively prune visual features relevant to question tokens, enhancing model efficiency. We validate our method across multiple video benchmarks, which demonstrate that PruneVid can prune over 80\% tokens while maintaining competitive performance combined with different model networks. This highlights its superior effectiveness and efficiency compared to existing pruning methods. Code: \url{https://github.com/Visual-AI/PruneVid}

\end{abstract}

\section{Introduction}
\label{sec:intro}
Large Language Models (LLMs)~\citep{gpt4,qwen, llama} have significantly advanced multi-modal understanding owing to their exceptional reasoning capabilities and proficiency in following instructions. Within the realm of video understanding, recent studies~\citep{llama-vid, videollava, videollama, videochat2, videochat, pllava, tarsier} have capitalized on the use of pre-trained LLMs as foundational models to address video question-answering tasks. However, the redundancy inherent in video content can lead to significant computational expenses for LLMs due to the quadratic complexity of attention mechanisms.  Consequently, effectively reducing the number of video tokens while preserving model performance emerges as an intriguing area of research.

Previous approaches attempt to address this challenge in various ways. LLaMA-VID~\citep{llama-vid} proposes compressing each frame into two distinct tokens: context and content tokens. However, this method necessitates extensive pretraining and fine-tuning phases, which limits its broader applicability with readily available video LLMs. Alternatively, LLaVA-PruMerge \citep{llavaprumerge} leverages the correlation between the \texttt{[CLS]} token and patch tokens within CLIP~\citep{clip} to identify important visual tokens and merges other less important ones. Yet, this approach does not consider the relevance of the selected tokens to the questions being asked, potentially selecting tokens that are unrelated to the task at hand. In a related vein, methods like Look-M~\citep{look-m} and Elastic Cache~\citep{elasticcache} employ Key-Value (KV) cache eviction strategies~\citep{h2o, scissorhands} to merge the KV cache for multi-modal inputs. These strategies prioritize retaining text tokens or treating visual and textual tokens equally without explicitly identifying the informative visual tokens. Moreover, eviction-based methods require encoding all visual tokens during the prefilling stage, which becomes inefficient when handling long visual sequences. Recently, FastV \citep{fastv} has leveraged attention patterns in LLMs to prune visual tokens. However, it is not specifically tailored for video understanding and does not adequately address the reduction of video inputs. 

\begin{figure}[t]
    \centering
    \includegraphics[width=\linewidth]{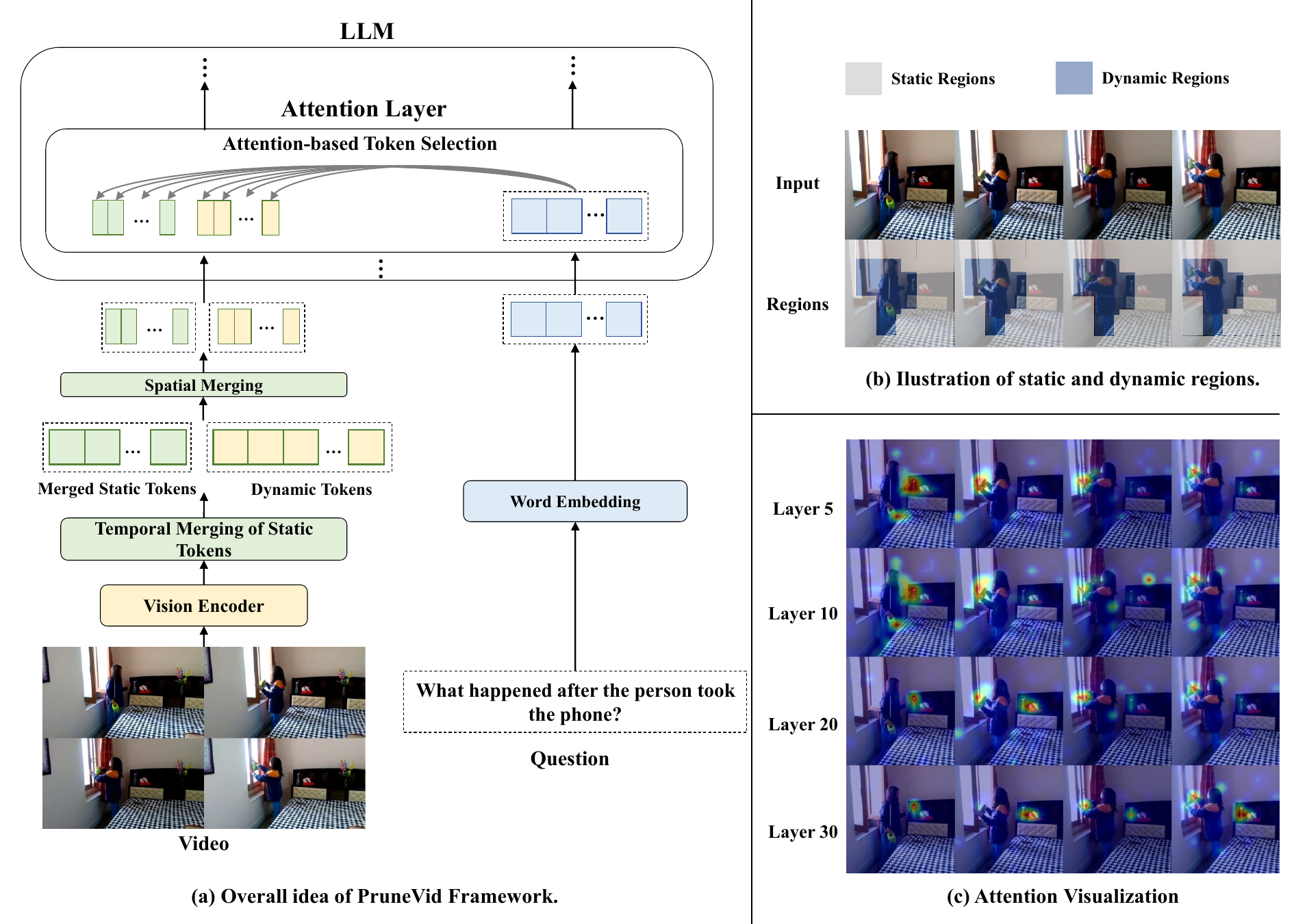}
    \caption{(a) PruneVid first identifies the static regions in the video that exhibit minimal variation, thereby compressing the redundancy of static tokens along the temporal dimension. It then further reduces spatial redundancy through compression in the spatial dimension. Subsequently, within the LLM, PruneVid utilizes question-to-visual attention scores to guide the selection of relevant visual tokens. (b) Static regions refer to areas with minimal change, while dynamic regions exhibit motion. Therefore, static regions can be compressed together along the temporal dimension. (c) Visualization of how attention evolves from shallow to deep layers (32 layers in total). The question tokens attend to semantically related visual regions (\emph{e.g.}, the hands and window) throughout different layers.}
    \label{fig:teaser_image}
\end{figure}

Building on the analysis presented above, we identify three essential criteria that an optimal pruning method for multi-modal video understanding ought to meet: (1) It should ideally be training-free, facilitating smooth integration with readily available models while reducing the need for extensive retraining or fine-tuning. (2) Inherent video redundancy needs to be reduced to save computations on tokens with similar representations along both spatial and temporal dimensions. (3) It is crucial to retain visual tokens specifically relevant to the given questions. This ensures the model maintains high performance while efficient and mitigates the risk of hallucinations when LLMs lack pertinent information~\citep{visualhallucination}.

To achieve our objectives, we present PruneVid, a training-free approach for pruning video tokens to achieve efficient video understanding. As shown in ~\cref{fig:teaser_image} (a), our method initially identifies static regions with minimal variation due to motion or camera movements, which could be interpreted as background (illustrated in \cref{fig:teaser_image} (b)). We merge these static tokens along the temporal dimension to reduce the computational burden from redundant temporal data. Next, we employ a clustering technique~\citep{DPC-KNN} to merge similar spatial tokens for both static and dynamic regions. In the subsequent step within the LLM, we use attention scores between the question and video tokens in an intermediate layer to discern and preserve the discriminative visual tokens essential for answering the question, while pruning irrelevant ones. As depicted in \cref{fig:teaser_image} (c), attention visualizations consistently highlight crucial features, such as hand movements and related objects (e.g., a window), which are directly relevant to the question. This indicates that important visual regions can be effectively pinpointed using attention layers, benefiting from the LLM’s reasoning and instruction-following prowess. Additionally, for the KV caches from previous layers, we retain the essential visual tokens and eliminate others, thus reducing computational demands during the decoding phase.

We integrate PruneVid with three video LLMs: PLLaVA~\citep{pllava}, ST-LLM~\citep{stllm}, and LLaVA-OneVision~\citep{llava-onevision}, and evaluate their performance on several video benchmarks, including MVBench~\citep{videochat2}, Video-MME~\citep{videomme}, Egoschema~\citep{egoschema}, and VideoChatGPT-Bench~\citep{videochatgpt}. Our extensive experiments demonstrate that PruneVid can prune over 80\% of visual tokens with only minimal performance degradation in certain cases. Notably, our method can occasionally enhance model performance. Furthermore, it achieves competitive results compared to the baseline model while boosting inference speed—up to 1.55 times faster—reducing FLOPs by 74\% to 80\%, and minimizing GPU memory usage.

The main contributions of this paper are as follows:
(1) We introduce PruneVid, a framework that efficiently prunes video tokens for video understanding without the need for retraining or fine-tuning, which can be seamlessly integrated with off-the-shelf video LLMs.
(2) We introduce a token pruning method that minimizes video redundancy by merging static tokens over time and clustering spatially similar ones. Furthermore, our approach leverages attention scores between the question and video tokens within the LLM to retain only the visual tokens pertinent to answering the questions.
(3) Extensive experiments are conducted across multiple benchmarks to demonstrate that PruneVid can consistently achieve superior efficiency and effectiveness with different video LLMs compared to existing approaches.

\section{Related Work}
\label{sec:related work}
\subsection{Video Large Language Model}
Recent advancements in Video LLMs focus on enabling LLMs to comprehend video content. These approaches are broadly categorized into training-free methods and training-required methods.

For training-free approaches~\citep{wu2024freeva,igvlm,sfllava}, they directly adapt the image LLMs for video tasks. FreeVA~\citep{wu2024freeva} compacts frame features for LLM processing, and IG-VLM~\citep{igvlm} merges frames into a single grid, simplifying video-to-image conversion. SF-LLaVA~\citep{sfllava} uses a SlowFast~\citep{slowfast} network design, balancing detailed spatial analysis with broad temporal scope efficiently within existing LLM token limits. These methods are ingeniously simple but are limited to handling only brief video clips due to their reliance on the inherent abilities of LLMs to understand temporal sequences.

Conversely, training-required Video LLMs improve comprehension by using extensive video datasets. Models like Video-ChatGPT~\citep{videochatgpt}, Video-LLaVA~\citep{videollava}, and PLLaVA~\citep{pllava} extend Image LLMs with video-specific tuning, greatly enhancing complex video understanding. Other approaches, such as VideoChat2~\citep{videochat2}, VILA~\citep{vila}, Tarsier~\cite{tarsier}, Chat-UniVi~\citep{chatunivi}, LLaMA-VID~\citep{llama-vid}, and ST-LLM~\cite{stllm}, optimize token usage, refine training protocols, advance vision-audio integration, or use dynamic masking, thereby advancing video content analysis. Recently, LLaVA-OneVision~\citep{llava-onevision} expanded the LLaVA~\citep{llava} architecture to incorporate more visual signals, achieving strong performance on video benchmarks.

Unlike the methods mentioned above, PruneVid aims to enhance the efficiency of existing video LLMs without additional training, which can be applied to both training-free and training-required methods. 

\subsection{Visual Token Pruning}
Due to the quadratic computational complexity inherent in attention mechanisms, optimizing efficiency through token pruning becomes essential. This optimization highlights a crucial distinction between methods designed for vision-centric and multi-modal tasks.

DynamicViT~\citep{rao2021dynamicvit} employs a prediction module to selectively prune less important tokens, thereby streamlining the model’s efficiency in processing visual data. Similarly, FastViT~\citep{vasu2023fastvit} reduces architectural complexity and memory demands through a novel token mixing operation, catering specifically to vision-only models. Further contributing to this domain, Token Merging (ToMe)~\citep{tokenmerge} merges tokens via token matching, while SPViT~\citep{spvit} introduces a method for softly aggregating redundant tokens into a single ‘package token’, efficiently preserving essential information while minimizing computational load.

Shifting away from vision-centric methods, LLaVA-Prumerge focuses on pruning visual tokens in multi-modal tasks. By employing an adaptive token reduction strategy, which utilizes CLIP’s inherent attention characteristics~\citep{clip}, LLaVA-Prumerge improves multi-modal understanding efficiency. FastV~\cite{fastv} addresses the inefficiency of visual attention patterns by pruning visual tokens with lower attention weights relative to the \texttt{[EOS]} token.


However, previous methods lack a specific focus on multi-modal video understanding. In contrast, this paper presents PruneVid, designed to eliminate redundancy in videos while utilizing LLMs to identify relevant video tokens for question answering. This approach enhances model efficiency without compromising performance.

\section{Method}
\label{sec:method}

Our method is designed to efficiently process video data by minimizing redundancy in visual tokens before inputting them into the LLM and identifying question-relevant visual tokens within the LLM. In this section, we introduce the necessary preliminaries and provide a detailed explanation of our method.

\subsection{Preliminaries}

\subsubsection{Pre-filling Stage}

In the pre-filling stage, the model processes the input question tokens and visual tokens to construct the initial representations and prepare the key-value (KV) caches for attention computations. Let $\bm{X}_q \in \mathbb{R}^{N_q \times C}$ denote the question tokens, where $N_q$ is the length of the question and $C$ is the channel dimension.

After the token merging, we obtain a compressed set of visual tokens $\tilde{\bm{X}}_v \in \mathbb{R}^{N_v' \times C}$, where $N_v'$ is the reduced number of visual tokens. The combined input sequence $\bm{X} \in \mathbb{R}^{(N_q + N_v') \times C}$ is formed by concatenating the question tokens and the merged visual tokens. The model employs a Transformer architecture with $L$ layers. In each layer $l$, the self-attention mechanism computes queries $\bm{Q}^{(l)}$, keys $\bm{K}^{(l)}$, and values $\bm{V}^{(l)}$ through linear projections of the input.

Based on this, the attention scores are computed using scaled dot-product attention with causal masking to prevent attending to future positions:
\begin{equation}
    \label{eq:attention_formula}
    \bm{A}^{(l)} = \text{Softmax} \left( \frac{\bm{Q}^{(l)} (\bm{K}^{(l)})^\top}{\sqrt{C}} + \bm{m} \right),
\end{equation}
where $\bm{m} \in \mathbb{R}^{(N_q + N_v') \times (N_q + N_v')}$ is a causal mask with entries $m_{ij} = -\infty$ if position $i < j$ (future positions) and $0$ otherwise.

The KV caches $\bm{KV}^{(l)} = (\bm{K}^{(l)}, \bm{V}^{(l)})$ are stored for each layer $l$ to facilitate efficient computation during decoding.

\subsubsection{Decoding Stage}

In the decoding stage, the model generates the answer tokens autoregressively, utilizing the stored KV caches from the pre-filling stage. At each decoding step, given the previously generated tokens, the model computes the necessary representations to predict the next token. By using the KV caches, the model efficiently attends to the input sequence without recomputing the attention for the entire sequence. This process reduces computational overhead and speeds up the generation of the response.

\begin{figure}[t]
    \centering
    \includegraphics[width=0.9\linewidth]{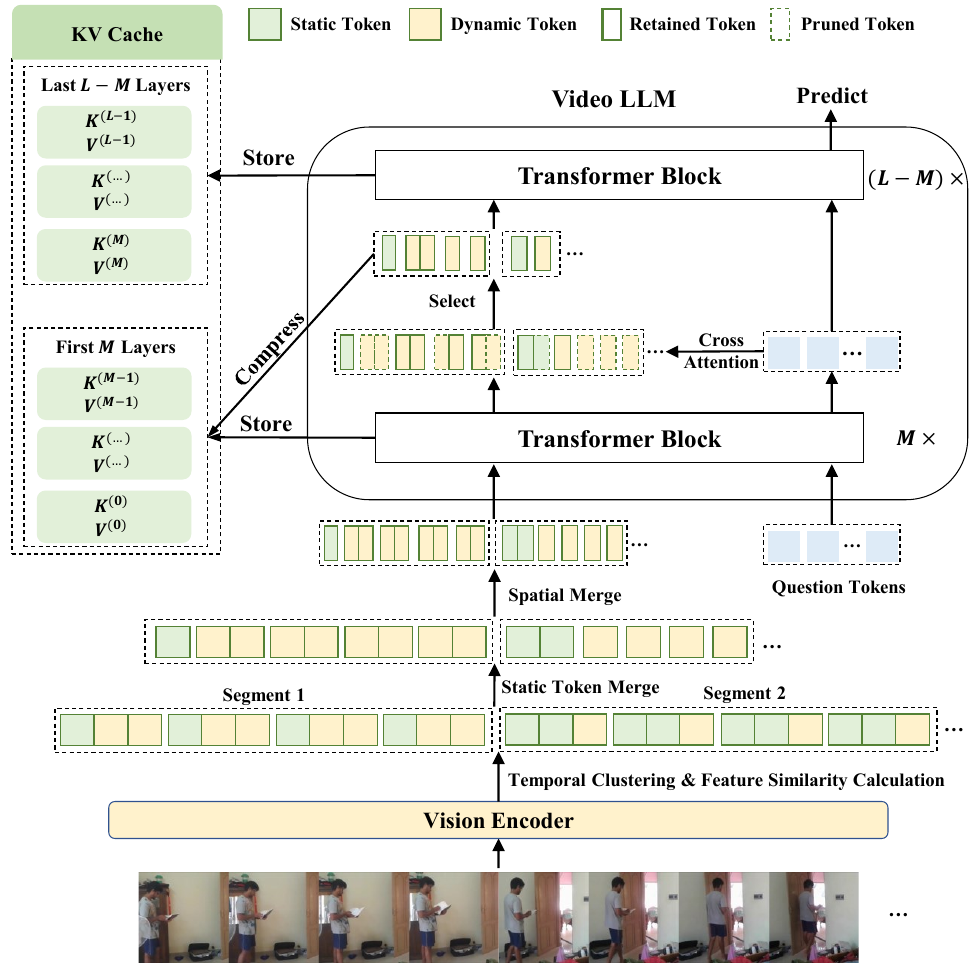}
    \caption{Illustration of the PruneVid framework. We begin by segmenting the video into different scenes and then decouple the video tokens into static and dynamic ones. Next, we compress the static tokens along the temporal dimension and merge similar tokens in the spatial dimension to further reduce redundancy. Afterward, by using the question-to-video attention weights learned from an intermediate layer, we determine which tokens should be pruned to improve efficiency.}
    \label{fig:method}
\end{figure}

\subsection{Spatial-Temporal Token Merging}

As depicted in \cref{fig:method}, given an input video consisting of $T$ frames, we first extract visual tokens from each frame using a visual encoder. Let $\bm{X}_v^{(t)} \in \mathbb{R}^{N_v \times C}$ denote the visual tokens for frame $t$, where $N_v$ is the number of tokens per frame. The complete set of visual tokens for the video is then $\bm{X}_v = \{\bm{X}_v^{(1)}, \bm{X}_v^{(2)}, \dots, \bm{X}_v^{(T)}\}$.

To identify temporal segments with different scenes in the video, we perform temporal clustering based on the visual content. We compute the average pooled feature $\bm{f}^{(t)} \in \mathbb{R}^{C}$ for each frame $t$ by averaging over its tokens. Using the features $\{\bm{f}^{(1)}, \dots, \bm{f}^{(T)}\}$, we employ the Density Peaks Clustering with $k$-Nearest Neighbors (DPC-KNN)~\citep{DPC-KNN} algorithm to group the frames into $B$ temporal segments $\{\mathcal{T}_1, \mathcal{T}_2, \dots, \mathcal{T}_{B}\}$. Here, $B$ is linearly related to $T$ with the coefficient $\gamma$, and each segment $\mathcal{T}_b$ comprises a subset of consecutive frames with similar content.

Within each temporal segment $\mathcal{T}_b$, we analyze the spatial tokens across the frames to identify static tokens—tokens that remain largely unchanged throughout the segment. For each spatial location $i$ (where $1 \leq i \leq N_v$), we extract the sequence of tokens $\{\bm{X}_v^{(t)}(i) \mid t \in \mathcal{T}_b\}$ and compute the feature similarities between every pair of tokens in this sequence. Specifically, for tokens at times $t$ and $t'$ within $\mathcal{T}_b$, the similarity is measured using cosine similarity $s_{i}^{(t,t')}$:
\begin{equation}
    s_{i}^{(t,t')} = \frac{\bm{X}_v^{(t)}(i)^\top \bm{X}_v^{(t')}(i)}{\| \bm{X}_v^{(t)}(i) \| \| \bm{X}_v^{(t')}(i) \|}.
\end{equation}

We then compute the average similarity for each spatial location $i$ within the segment:
\begin{equation}
    \bar{s}_i = \frac{2}{|\mathcal{T}_b|(|\mathcal{T}_b| - 1)} \sum_{t, t' \in \mathcal{T}_b, t < t'} s_{i}^{(t,t')}.
\end{equation}

Tokens with average similarity above a threshold $\tau$ are considered static:
\begin{equation}
    \mathcal{I}_\text{static} = \left\{ i \mid \bar{s}_i \geq \tau \right\}.
\end{equation}

For these static tokens, we perform temporal averaging within the segment to compress temporal redundancy:
\begin{equation}
    \tilde{\bm{X}}_v^{(b)}(i) = \frac{1}{|\mathcal{T}_b|} \sum_{t \in \mathcal{T}_b} \bm{X}_v^{(t)}(i), \quad \forall i \in \mathcal{I}_\text{static}.
\end{equation}

The dynamic tokens, corresponding to $\mathcal{I}_\text{dynamic} = \{1, \dots, N_v\} \setminus \mathcal{I}_\text{static}$, are retained without temporal averaging.

To further reduce spatial redundancy, we perform spatial clustering within each frame also using the DPC-KNN algorithm. We apply this clustering separately to the static and dynamic tokens. For frame $t$, we cluster the tokens $\{\bm{X}_v^{(t)}(i) \mid i \in \mathcal{I}_\text{static}\}$ and $\{\bm{X}_v^{(t)}(i) \mid i \in \mathcal{I}_\text{dynamic}\}$ to obtain clusters $\mathcal{C}_1^{(t)}, \dots, \mathcal{C}_{C_s}^{(t)}$ for the static tokens and $\mathcal{D}_1^{(t)}, \dots, \mathcal{D}_{C_d}^{(t)}$ for the dynamic tokens. For both static and dynamic tokens, the cluster number is linearly related to $|\mathcal{I}_\text{static}|$ and $|\mathcal{I}_\text{dynamic}|$ with the coefficient $\beta$. We average the tokens within each cluster to represent them with a single token $\tilde{\bm{X}}_v^{(t)}(c)$ for $c = 1, \dots, C_s$ in the static clusters, and similarly for the dynamic clusters $\tilde{\bm{X}}_v^{(t)}(d)$ for $d = 1, \dots, C_d$.

After these merging operations, we obtain a reduced set of visual tokens $\tilde{\bm{X}}_v$ with significantly less redundancy.
The merged visual tokens for the entire video are then collected and concatenated to form the final token sequence to be input to the LLM:
\begin{equation}
    \tilde{\bm{X}}_v = \bigcup_{b=1}^{B} \left( \tilde{\bm{X}}_v^{(b)} \right),
\end{equation}
where $\tilde{\bm{X}}_v^{(b)}$ contains the merged tokens from segment $\mathcal{T}_b$.

\subsection{LLM-Guided Token Selection}

We further reduce the visual tokens by leveraging the LLM's internal attentions to select the most relevant tokens with respect to the given question.

Consider the LLM with $L$ layers. During the pre-filling stage, we target the $M$-th layer, where $1 \leq M \leq L$, to compute cross-attention weights between the question tokens and the merged visual tokens to obtain a measure of relevance.

At the $M$-th layer, we calculate the attention scores $\bm{A}^{(M)} \in \mathbb{R}^{(N_q + N_v') \times (N_q + N_v')}$ according to \cref{eq:attention_formula}. To obtain the cross-attention scores between question and visual tokens, we extract a submatrix $\bm{A}_{qv}^{(M)} \in \mathbb{R}^{N_q \times N_v'}$ as follows:
\begin{equation}
    \bm{A}_{qv}^{(M)} = \bm{A}^{(M)}[N_v':, :N_v'],
\end{equation}
where $\bm{A}^{(M)}[N_v':, :N_v']$ selects the attention scores from the question tokens to the visual tokens.

Next, we compute the maximum attention values $\bm{a}_v \in \mathbb{R}^{N_v'}$ for each visual token by applying max pooling over all question tokens. This approach captures the most informative tokens, as not all question tokens are equally important:

$$ \bm{a}_v = \max_{i=1}^{N_q} \bm{A}_{qv}^{(M)}(i, :). $$

We then sort the attention scores in descending order and select the top $\alpha\%$ of visual tokens. The set of indices for the selected tokens is represented by $\mathcal{S}$ and is defined as:
\begin{equation}
    \mathcal{S} = \left\{ j \in \{1, \dots, N_v'\} \mid \text{Rank} \left( \bm{a}_v(j) \right) \leq \left\lceil \alpha N_v' \right\rceil \right\},
\end{equation}
where $\text{Rank} \left( \bm{a}_v(j) \right)$ indicates the rank of $\bm{a}_v(j)$ within the sorted attention scores, and $\left\lceil \cdot \right\rceil$ denotes the ceiling function.

By focusing on the top $\alpha\%$ tokens, we align the model's attention with the most question-relevant visual information. To finalize the pre-filling stage, we combine the selected visual tokens with the question tokens, enabling processing in the remaining $(L - M)$ layers of the LLM. The KV vectors derived from the retained visual tokens and question tokens, calculated in the last $(L - M)$ layers, are stored in the KV cache for the decoding process.

\subsubsection{Compressed Key-Value Caches}

To reduce memory and computational costs during the decoding stage, we compress the KV caches stored from the previous $M$ layers by retaining only the selected visual tokens. For each layer $l$ ($1 \leq l \leq M$), the original key and value matrices for the visual tokens are $\bm{K}_v^{(l)} \in \mathbb{R}^{N_v' \times C}$ and $\bm{V}_v^{(l)} \in \mathbb{R}^{N_v' \times C}$.

We create the compressed key and value matrices $\tilde{\bm{K}}_v^{(l)}$ and $\tilde{\bm{V}}_v^{(l)}$ by selecting the rows corresponding to the indices in $\mathcal{S}$:
\begin{equation}
    \tilde{\bm{K}}_v^{(l)} = \bm{K}_v^{(l)}\left[ \mathcal{S}, : \right], \quad \tilde{\bm{V}}_v^{(l)} = \bm{V}_v^{(l)}\left[ \mathcal{S}, : \right],
\end{equation}
where $\bm{K}_v^{(l)}\left[ \mathcal{S}, : \right]$ and $\bm{V}_v^{(l)}\left[ \mathcal{S}, : \right]$ denote the selection of rows corresponding to the indices in $\mathcal{S}$.

Similarly, we adjust the key and value matrices for the entire sequence by combining the question tokens and the selected visual tokens:
\begin{equation}
    \tilde{\bm{K}}^{(l)} = \left[ \tilde{\bm{K}}_v^{(l)} ; \bm{K}_q^{(l)} \right], \quad \tilde{\bm{V}}^{(l)} = \left[ \tilde{\bm{V}}_v^{(l)} ; \bm{V}_q^{(l)} \right],
\end{equation}
where $\bm{K}_q^{(l)}$ and $\bm{V}_q^{(l)}$ are the key and value matrices for the question tokens.
By compressing the KV caches, we effectively reduce the sequence length from $N_q + N_v'$ to $N_q + |\mathcal{S}|$, where $|\mathcal{S}|$ represents the total number of selected visual tokens.

This compression significantly reduces the memory requirements and computational complexity during decoding, enabling efficient processing of long video sequences within the LLM framework.

\section{Experiment}
\label{sec:exp}
\subsection{Datasets and Evaluation Metrics}
\textbf{Generic Multi-Choice VideoQA.} MVbench~\citep{videochat2} encompasses 20 temporally challenging tasks that cannot be addressed using a single frame. Each task includes 200 test samples, formatted as multiple-choice VideoQA. These samples require the model to choose the correct answer from several provided options.

\textbf{Long-form Multi-Choice VideoQA.} We conduct evaluations of our models using two well-regarded benchmarks for long-form video benchmarks: Video-MME \citep{videomme} and Egoschema \citep{egoschema}. In these evaluations, the models are tasked with selecting the correct answer from multiple-choice options.

\textbf{Text Generation.} VideoChatGPT-Bench, introduced by~\citep{videochatgpt}, focuses on five aspects: Correctness of Information (CI), Detail Orientation (DO), Contextual Understanding (CU), Temporal Understanding (TU), and Consistency (CO). For evaluation, we use \texttt{GPT-3.5-Turbo-0125} for scoring.

\subsection{Baselines}

To evaluate the effectiveness of our approach, we compare it with three visual token pruning methods: LLaVA-PruMerge~\citep{llavaprumerge}, Look-M~\citep{look-m}, and FastV~\citep{fastv}. LLaVA-PruMerge utilizes attention score sparsification within CLIP to identify crucial tokens and employs an outlier detection method to adaptively determine the optimal pruning ratio. Conversely, Look-M extends the concept of text-only KV cache compression to a multi-modal context by implementing strategies for evicting text-prior KV pairs and merging them through a pivotal merging strategy. Additionally, FastV~\citep{fastv} uses attention weights to prune visual tokens with low attention scores. To ensure a fair comparison, we use the official implementations of these methods and apply them to video benchmarks.

\subsection{Implementation Details}
All experiments are conducted on NVIDIA A100 GPUs with 80GB of memory. We implement PruneVid, LLaVA-PruMerge, Look-M, and FastV on three video LLMs: PLLaVA~\citep{pllava}, ST-LLM~\citep{tarsier}, and LLaVA-OneVision~\citep{llava-onevision}. LLaVA-PruMerge is incompatible with ST-LLM, so it is excluded from related comparisons. As per the official settings, the input frames are set to 16 for both PLLaVA and ST-LLM, and 32 for LLaVA-OneVision. For the VideoChatGPT-Bench, ST-LLM uses 64 input frames. Besides, The threshold $\tau$ is set to 0.8, the temporal segment ratio $\gamma$ is 0.25, and the cluster ratio $\beta$ is 0.5. Across all benchmarks, the token selection ratio $\alpha$ is 0.4, and attention calculations use the 10th layer ($M$). For FastV, we prune the tokens at the 2nd layer and set the retained ratio to 0.3 to achieve roughly comparable FLOPs to our method. Additionally, the FLOPs in the experiments are measured in relation to the visual tokens in the LLM.


\begin{table}[t]
\centering
\setlength{\tabcolsep}{2pt} 
\scriptsize 
\caption{Performance and efficiency comparison across different methods and benchmarks. The best results of pruning methods are \textbf{bolded}.}
\begin{tabular}{@{}lccccc|cccccc@{}}
\toprule
\multirow{2}{*}{\textbf{Method}} & \multirow{2}{*}{\textbf{Retained Ratio}} & \multirow{2}{*}{\textbf{FLOPs ($\times$)}} & \multirow{2}{*}{\textbf{MVBench}} & \multirow{2}{*}{\textbf{VideoMME}} & \multicolumn{1}{c}{\textbf{EgoSchema}} & \multicolumn{6}{c}{\textbf{VideoChatGPT-Bench}} \\
\cmidrule(lr){6-6} \cmidrule(lr){7-12}
 & & & & & \textbf{Subset / Fullset} & \textbf{TU} & \textbf{CU} & \textbf{CO} & \textbf{DO} & \textbf{CI} & \textbf{Avg} \\
\midrule
PLLaVA                   & 100.0\% & 1.00$\times$ & 46.6  & 44.4  & 47.8 / 42.6 & 2.33 & 3.62 & 2.93 & 2.86 & 3.21 & 2.99 \\
PLLaVA w/ FastV          & 30.0\%  & 0.33$\times$ & 46.1  & 43.6  & 46.2 / 41.0 & 2.38 & 3.49 & 2.89 & 2.76 & 3.14 & 2.93 \\
PLLaVA w/ Prumerge       & 55.7\%  & 0.53$\times$ & 45.6  & 43.8  & 45.2 / 40.4 & 2.34 & 3.52 & 2.90 & 2.76 & 3.15 & 2.93 \\
PLLaVA w/ Look-M         & 20.0\%  & $1.00\times$ & 46.6  & 44.3  & 47.0 / 42.3 & 2.28 & 3.41 & 2.75 & 2.65 & 3.00 & 2.82 \\
\rowcolor[HTML]{E3F2FD} 
PLLaVA w/ Ours           & \textbf{16.2\%}  & \textbf{0.23$\times$} & \textbf{47.6}  & \textbf{45.0}  & \textbf{49.0} / \textbf{42.6} & \textbf{2.44} & \textbf{3.51} & \textbf{2.99} & \textbf{2.78} & \textbf{3.20} & \textbf{2.98} \\
\midrule
ST-LLM                   & 100.0\% & 1.00$\times$ & 54.9  & 42.0  & 56.2 / 45.6 & 2.46 & 3.46 & 2.66 & 2.63 & 3.08 & 2.86 \\
ST-LLM w/ FastV          & 30.0\%  & 0.37$\times$ & 42.9  & 34.5  & 48.0 / 38.5 & 2.01 & 2.23 & 1.55 & 1.94 & 1.69 & 1.88 \\
ST-LLM w/ Look-M         & 20.0\%  & 1.00$\times$ & 54.0  & 40.6  & 54.0 / 44.5 & 2.35 & 3.41 & 2.60 & 2.51 & 3.01 & 2.78 \\
\rowcolor[HTML]{E3F2FD} ST-LLM w/ Ours        & \textbf{15.1\%}  & \textbf{0.26$\times$} & \textbf{54.3}  & \textbf{41.4}  & \textbf{54.6} / \textbf{44.7} & \textbf{2.40} & \textbf{3.43} & \textbf{2.63} & \textbf{2.60} & \textbf{3.04} & \textbf{2.82} \\
\midrule
LLaVA-OneVision                 & 100.0\% & 1.00$\times$ & 58.0  & 58.2  & 62.0 / 60.0 & 2.75 & 3.70 & 3.39 & 2.97 & 3.50 & 3.26 \\
LLaVA-OneVision w/ FastV        & 30.0\%  & 0.30$\times$ & 57.2  & 57.6  & 62.6 / 60.0 & 2.65 & 3.61 & 3.28 & 2.85 & 3.39 & 3.16 \\
LLaVA-OneVision w/ Prumerge     & 55.2\%  & 0.49$\times$ & 52.9  & 56.7  & 62.2 / 60.0 & 2.72 & 3.64 & \textbf{3.32} & \textbf{2.94} & 3.44 & 3.21 \\
LLaVA-OneVision w/ Look-M       & 20.0\%  & 1.00$\times$ & 57.0  & 58.0  & 62.0 / \textbf{59.8} & 2.71 & 3.70 & 3.29 & 2.89 & 3.44 & 3.21 \\
\rowcolor[HTML]{E3F2FD} LLaVA-OneVision w/ Ours      & \textbf{17.0\%}  & \textbf{0.20$\times$} & \textbf{57.5}  & \textbf{58.6}  & \textbf{62.6} / 59.5 & \textbf{2.73} & \textbf{3.72} & 3.28 & \textbf{2.94} & \textbf{3.51} & \textbf{3.24} \\
\bottomrule
\end{tabular}
\label{tab:main result}
\end{table}

\subsection{Main Result}
As illustrated \cref{tab:main result}, our method consistently achieves the best performance in almost all cases compared to existing pruning methods (FastV, Prumerge, and Look-M) while retaining fewer tokens and achieving lower FLOPs. For instance, on PLLaVA, our approach retains only 16.2\% of tokens yet surpasses the performance of other pruning methods and even the baseline model under MVBench, VideoMME, and Egoschema. A similar pattern is observed for ST-LLM and LLaVA-OneVision, where our method maintains robust performance with retained ratios as low as 15.1\% and 17.0\%, respectively, across all benchmarks. This underscores the versatility of our approach in balancing accuracy with a substantial reduction in computational overhead.

Moreover, while Prumerge also maintains competitive accuracy on some models, it fails to do so with substantially reduced token budgets. Similarly, although Look-M can achieve decent performance, it requires using the vanilla attention implementation for all layers, resulting in relatively low efficiency. Additionally, we find that FastV struggles to maintain consistent performance across different models. For instance, while it performs well on PLLaVA and LLaVA-OneVision, its accuracy on ST-LLM is unsatisfactory, indicating a lack of robustness across diverse architectures. In contrast, our method effectively adapts by identifying and preserving only the most informative tokens for video understanding, thereby delivering strong overall performance with significantly reduced computational costs.

\begin{table}[t]
\caption{Efficiency comparison for visual token pruning methods. TTFT stands for time-to-first-token, which is commonly used for evaluating the efficiency of LLMs.}
    \centering
    \footnotesize
    \begin{tabular}{l|c|c|c|c}
        \toprule
        \textbf{Method} & \textbf{FLOPs ($\times$)} & \textbf{TTFT Speed Up ($\times$)} & \textbf{GPU Mem} & \textbf{Accuracy} \\
        \midrule
        Baseline & 1.00 $\times$ & 1.00 $\times$ & 20G & 46.6 \\
        Baseline w/FastV & 0.33 $\times$ & 1.15 $\times$ & 19G & 46.1 \\
        Baseline w/Prumerge & 0.53 $\times$ & 1.32 $\times$ & 19G & 45.6 \\
        Baseline w/Look-M & 1.00 $\times$ & 0.15 $\times$ & 35G & 46.6 \\
        Baseline w/Ours & \textbf{0.23 $\times$} & \textbf{1.55} $\times$ & \textbf{17G} & \textbf{47.6} \\
        \bottomrule
    \end{tabular}
\label{tab:efficiency_comparison}
\end{table}


\subsection{Diagnostic Study}
\label{sec:disagnostic}
In this section, we conduct a diagnostic study based on the PLLaVA model for efficiency comparison, hyper-parameter investigations, and qualitative visualizations.

\textbf{Efficiency Analysis.} As shown on \cref{tab:efficiency_comparison}, We compare multiple visual token pruning methods with respect to FLOPs, TTFT speed-up, GPU memory usage, and accuracy. Overall, FastV and Prumerge both demonstrate notable FLOPs reduction and moderate speed gains, though their influence on accuracy remains slightly adverse. By contrast, Look-M manages to preserve accuracy comparable to the baseline but incurs a very low TTFT speed up and escalates GPU memory to a high level, suggesting that its underlying layer-wise strategy increases overhead. In contrast, our proposed method achieves the most efficient balance by yielding the fastest TTFT speed up, the largest FLOPs reduction, the smallest memory footprint, and the highest accuracy, clearly underscoring the advantages of our token pruning approach over existing techniques.

\begin{figure}[t]
    \centering
    \includegraphics[width=\linewidth]{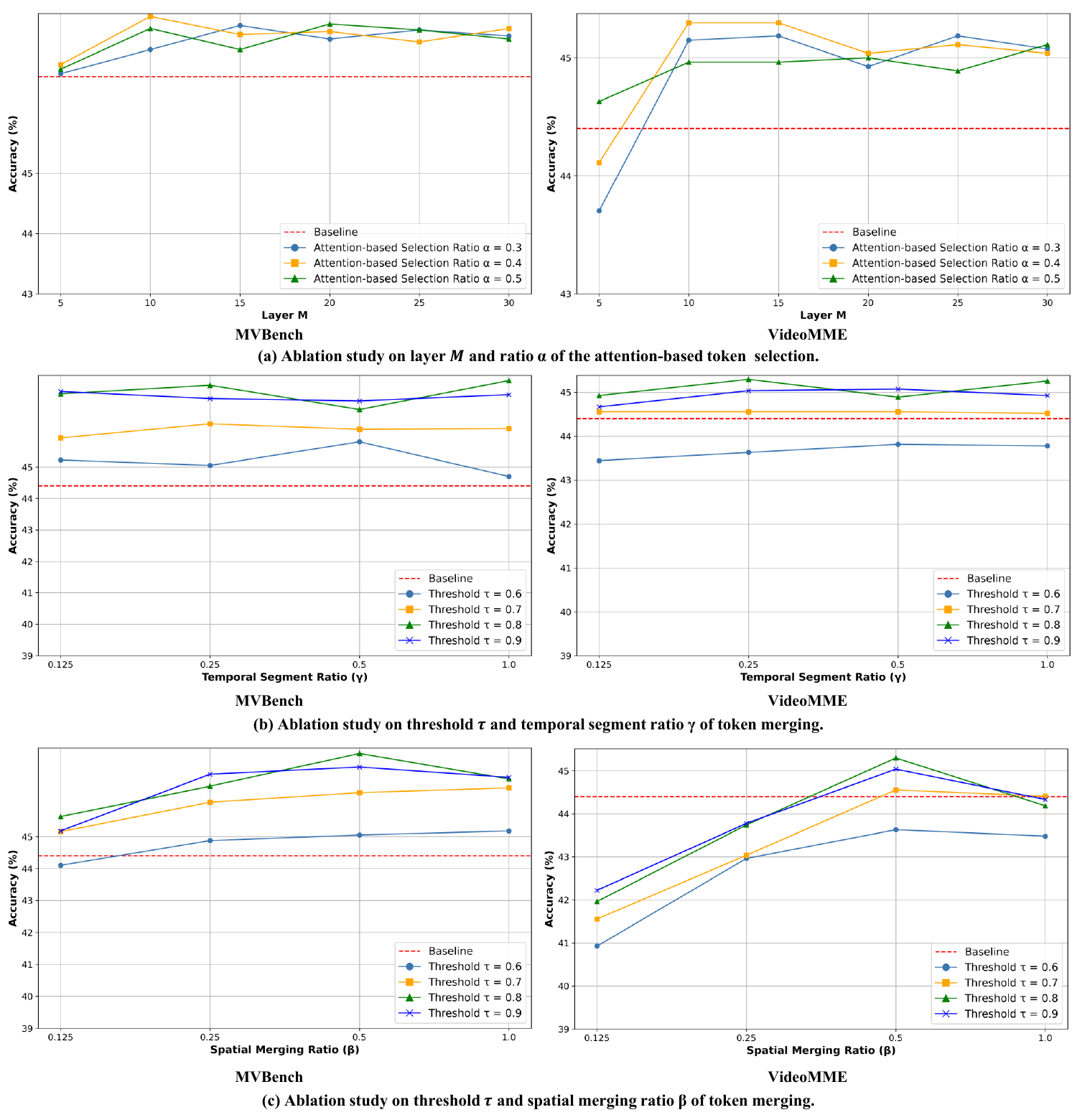}
    \caption{The ablation study of hyper-parameters.}
    \label{fig:ablation graphs}
\end{figure}

\begin{figure}[htbp]
    \centering
    \includegraphics[width=0.95\linewidth]{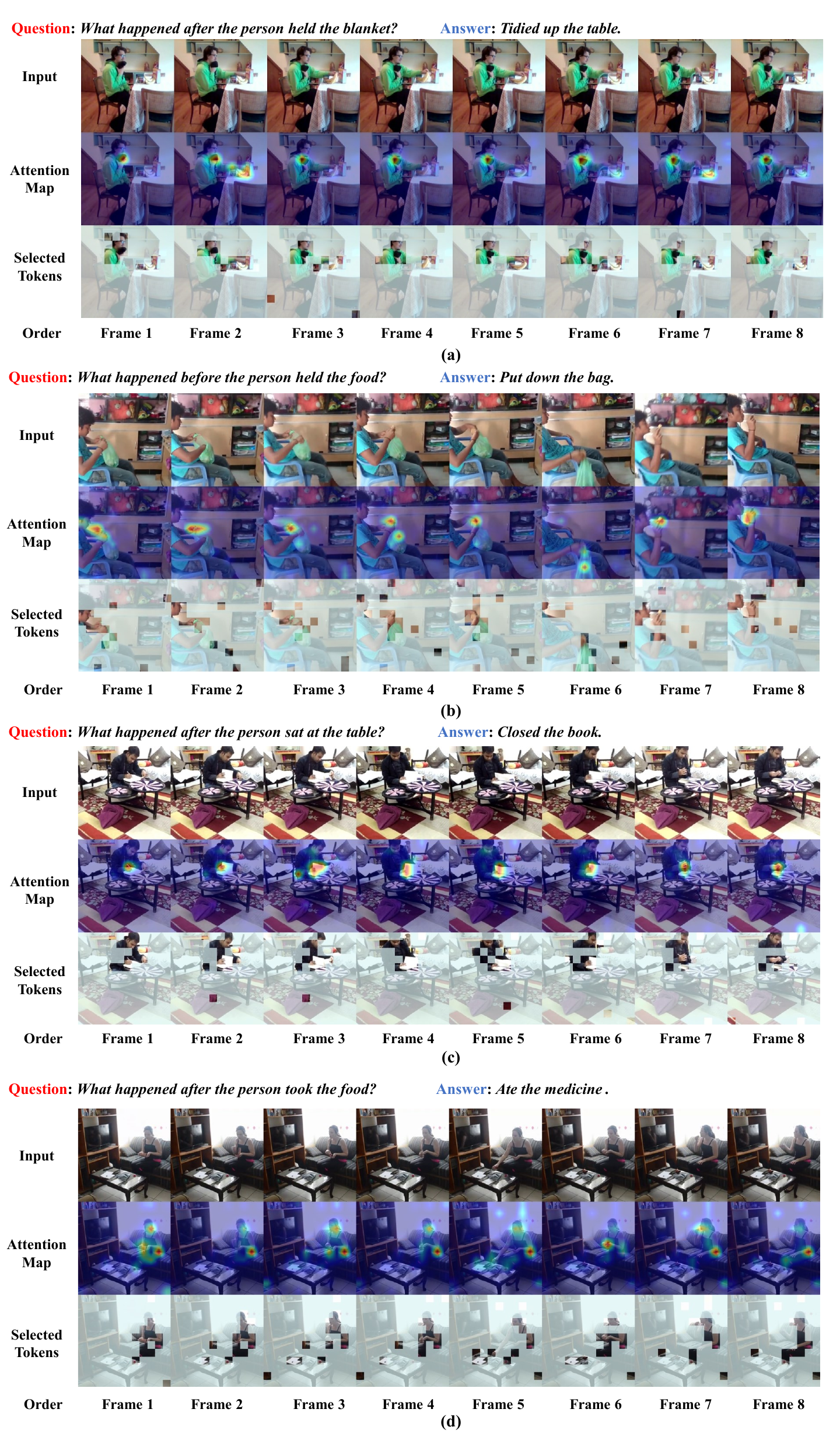}
    \caption{Visualization of the question-to-visual attentions and token selection of PruneVid.}
    \label{fig:selected tokens}
\end{figure}

\textbf{Ablation Study on Token Selection Ratio $\alpha$ and the Position of Pruning Layer $M$.} As shown in \cref{fig:ablation graphs} (a), we observe that when pruning attention from the 10th layer onward, model accuracy, despite minor fluctuations, gradually saturates. Therefore, selecting $M$ as 10 for token pruning results in lower computational costs compared to pruning at later layers. Additionally, we find that using a larger $\alpha$ does not necessarily yield better results. For instance, when $M$ is 10, an $\alpha$ of 0.4 achieves better accuracy than 0.5, as retaining more tokens may introduce irrelevant information, adversely affecting the outcome.

\textbf{Ablation Study on threshold $\tau$ and temporal segment ratio $\gamma$.} As depicted in \cref{fig:ablation graphs} (b), we observe that performance consistently improves as $\tau$ increases from 0.6 to 0.8, while the performance between $\tau=0.8$ and $\tau=0.9$ remains similar. Given that setting $\tau=0.8$ allows the model to merge more tokens along the temporal dimension, resulting in a higher compression ratio, we select $\tau=0.8$. Concerning the temporal segment ratio $\gamma$, the variation in its values does not significantly affect performance. We found that both $\gamma=0.25$ and $\gamma=1.0$ deliver good results across the two datasets. However, since $\gamma=1.0$ treats each input frame as an individual segment, which hinders effective temporal merging of static tokens, we choose to set $\gamma$ to 0.25.

\textbf{Ablation Study on threshold $\tau$ and spatial merging ratio $\beta$.} As shown in \cref{fig:ablation graphs} (c), the performance is comparable for $\tau$ values of 0.8 and 0.9, with $\tau=0.8$ being slightly superior, which is a similar phenomenon as in \cref{fig:ablation graphs} (b). Regarding the parameter $\beta$, setting it to 0.5 provides the optimal accuracy. This is because a smaller $\beta$ leads to overly aggressive merging of spatial tokens, which degrades performance. On the other hand, setting $\beta$ to 1.0 is unable to merge redundant tokens, which also adversely affects performance.

\textbf{Side-by-side Visualizations of Attention Maps and Token Selection.} In \cref{fig:selected tokens}, we present a side-by-side comparison demonstrating how our model selects tokens guided by attention scores, highlighting the LLM’s strength in focusing on informative regions related to the questions.

Below, we provide detailed analyses on how our model leverages the \textit{reasoning capabilities of LLMs} to locate relevant visual regions that are not explicitly mentioned in the questions:

\cref{fig:selected tokens} (a): The model identifies the key object mentioned in the question, the \textit{blanket}, and uses the temporal cue (\textit{after}) from the question to locate additional relevant visual elements, such as objects on the table and the person's hand movements in frames 2, 6, and 7. These details are not directly provided in the question and are inferred through the model's reasoning abilities.

\cref{fig:selected tokens} (b): The model accurately detects the action of the person holding food in frames 7 and 8, and infers that the presence of a \textit{bag} the person puts down is relevant for answering the question, even though the \textit{bag} is not mentioned. This demonstrates the model's ability to reason about relevant objects based on contextual cues.

\cref{fig:selected tokens} (c): Despite the absence of any mention of a \textit{book} in the question, the model correctly identifies critical visual regions related to the \textit{book} by reasoning over the visual content and context provided.

\cref{fig:selected tokens} (d): The model focuses on the person's hand movements, which are crucial for answering the question. Even though the question does not emphasize hand motions, the model infers the importance of these actions through reasoning.

These examples showcase how our model utilizes LLM reasoning to identify and focus on pertinent visual information that is not explicitly described in the questions.

\section{Conclusion}
We present PruneVid, a training-free visual token pruning method that enhances efficiency in multi-modal video understanding. By reducing video redundancy through merging static tokens over time and clustering similar spatial tokens, PruneVid minimizes the number of tokens processed. It leverages attention mechanisms within LLMs to retain only the visual tokens relevant to questions, ensuring high performance while reducing computational overhead. Experiments across multiple benchmarks demonstrate that PruneVid can prune over 80\% of visual tokens while maintaining, or even improving, model performance. By eliminating the need for retraining or fine-tuning, PruneVid offers a practical and efficient solution that integrates seamlessly with existing video LLMs.

\section*{Acknowledgments}
This work is supported by Hong Kong Research
Grant Council - Early Career Scheme (Grant No. 27208022), National Natural Science Foundation of China (Grant No. 62306251), and HKU Seed Fund for Basic Research.


\bibliography{iclr2025_conference}
\bibliographystyle{iclr2025_conference}

\clearpage
\appendix
\section{Appendix}
\label{appendix}
\begin{figure}[!h]
    \centering
    \includegraphics[width=\linewidth]{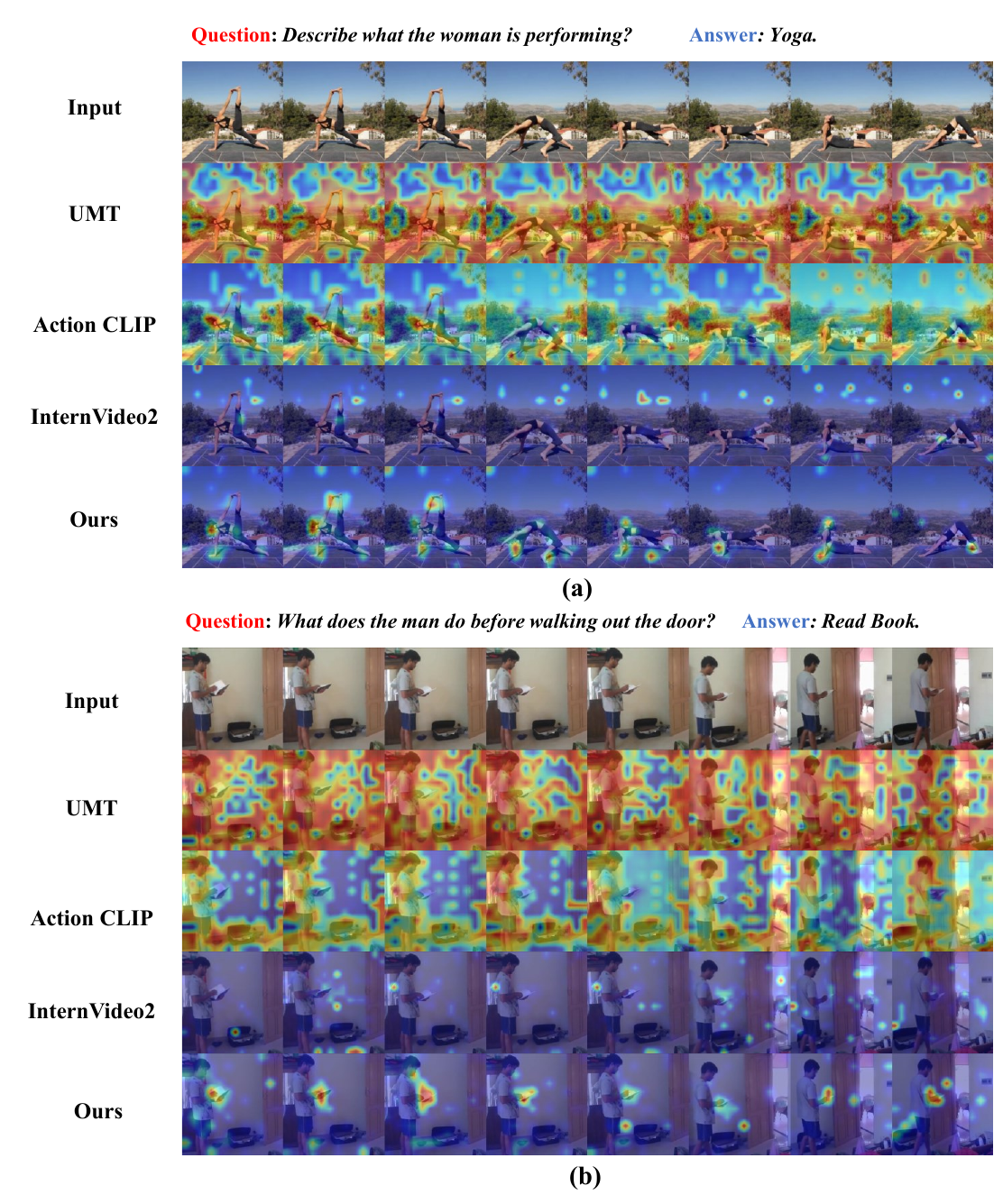}
    \caption{Attention map comparison of video encoders and our method.}
    \label{fig:attention1}
\end{figure}

\begin{figure}[!h]
    \centering
    \includegraphics[width=\linewidth]{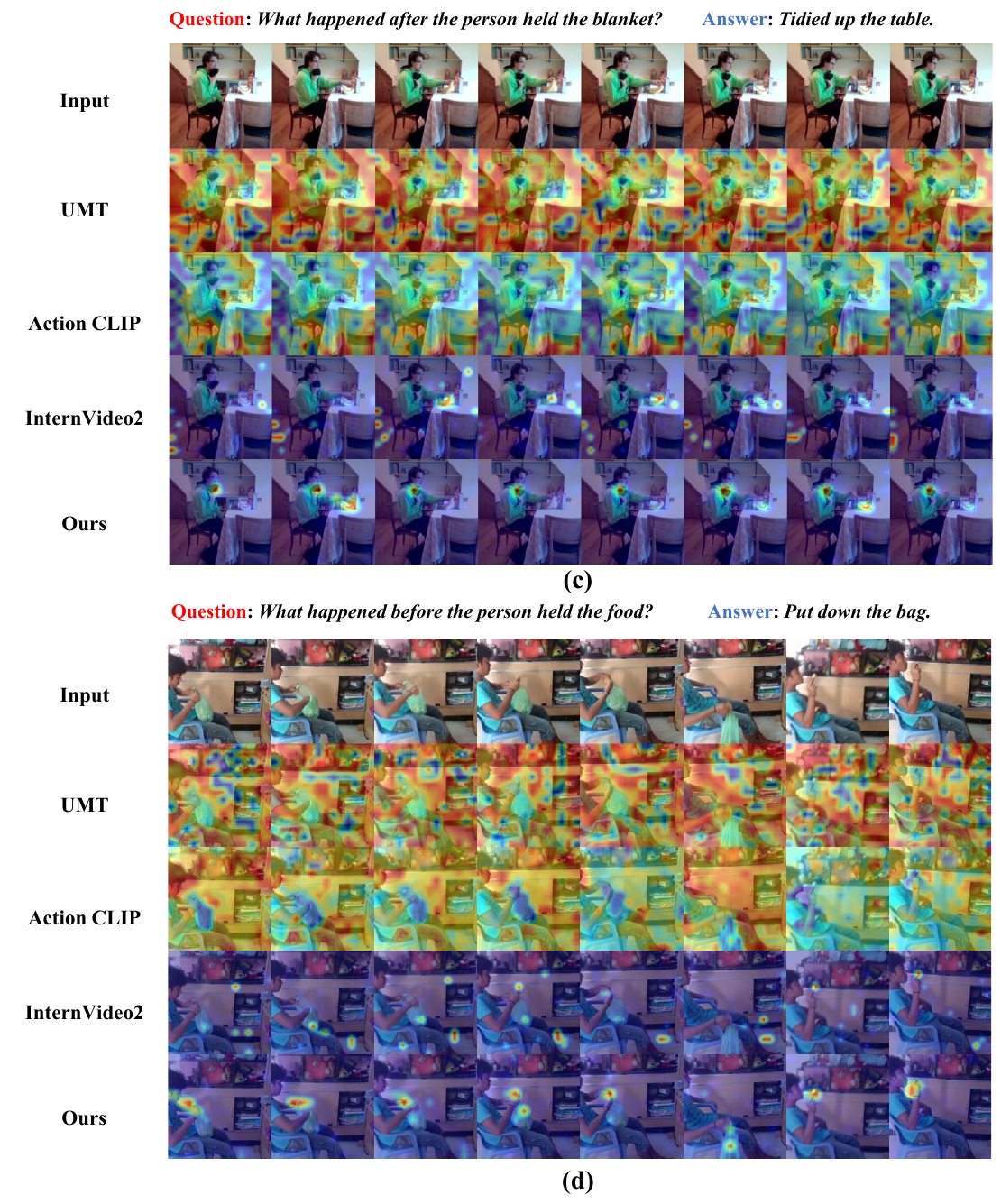}
    \caption{Attention map comparison of video encoders and our method.}
    \label{fig:attention2}
\end{figure}

\textbf{Attention Map Comparison.} In \cref{fig:attention1} and \cref{fig:attention2}, we include comparisons between our LLM’s attention maps and those of several strong video encoders, including UMT \citep{umt}, ActionCLIP \citep{wang2021actionclip}, and InternVideo2 \citep{wang2024internvideo2}. The results show that, unlike these models, the question-to-vision attentions in the LLM accurately focus on visual tokens that are pertinent to the question. In contrast, the other models often struggle to pinpoint key tokens and may focus on irrelevant objects or background elements. These observations suggest that LLMs possess a unique ability to align visual information with linguistic context through their reasoning capabilities, which is not simply a byproduct of standard attention mechanisms in typical video encoders.

\end{document}